\documentclass[review]{elsarticle}
\usepackage{amsmath}
\usepackage{setspace}
\usepackage{ulem}
\usepackage[perpage]{footmisc}
\usepackage{lineno}
\usepackage{booktabs}
\usepackage{subcaption}
\usepackage{algorithm,tabularx}
\usepackage[noend]{algpseudocode}
\usepackage{tikz}
\usepackage{color}
\usepackage{csquotes}
\usepackage{dsfont}
\usepackage{xhfill}
\usepackage{lineno}
\usepackage{soul}
\usepackage{emptypage}
\usepackage{rotating}
\usepackage{multirow}
\usepackage{booktabs}
\usepackage{makecell}
\usepackage{setspace}
\usepackage{parselines}
\usepackage{lineno}
\usepackage{setspace}
\usepackage{hyperref}

\modulolinenumbers[5]

\newlength\tbspace
\newcolumntype{L}{l<{\hspace{\tbspace}}}
\DeclareMathOperator*{\argminA}{arg\,min}

\makeatletter
\newcommand{\multiline}[1]{%
  \begin{tabularx}{\dimexpr\linewidth-\ALG@thistlm}[t]{@{}X@{}}
    #1
  \end{tabularx}
}
\makeatother
\algnewcommand{\do}{\textbf{do }}
\algnewcommand\Input{\item[\textbf{Input:}]}%
\algnewcommand\Output{\item[\textbf{Output:}]}%

\makeatletter
\def\ps@pprintTitle{%
 \let\@oddhead\@empty
 \let\@evenhead\@empty
 \def\@oddfoot{\centerline{\thepage}}%
 \let\@evenfoot\@oddfoot}
\makeatother

\PassOptionsToPackage{draft}{graphicx}

\begin{document}

\begin{frontmatter}

\title{Online Visual Tracking with One-Shot Context-Aware Domain Adaptation}
\author{Hossein Kashiani\corref{mycorrespondingauthor}}
\cortext[mycorrespondingauthor]{Corresponding author}
\ead{hossein\_kashiyani@alumni.iust.ac.ir}
\author{Amir Abbas Hamidi Imani}
\ead{hamidi\_a@elec.iust.ac.ir}
\author{Shahriar Baradaran Shokouhi}
\ead{bshokouhi@iust.ac.ir}
\author{Ahmad Ayatollahi}
\ead{ayatollahi@iust.ac.ir}

\address{School of Electrical Engineering, Iran University of Science and Technology, Tehran, Iran}

\begin{abstract}

Online learning policy makes visual trackers more robust against different distortions through learning domain-specific cues. However, the trackers adopting this policy fail to fully leverage the discriminative context of the background areas. Moreover, owing to the lack of sufficient data at each time step, the online learning approach can also make the trackers prone to over-fitting to the background regions. In this paper, we propose a domain adaptation approach to strengthen the contributions of the semantic background context. The domain adaptation approach is backboned with only an off-the-shelf deep model. The strength of the proposed approach comes from its discriminative ability to handle severe occlusion and background clutter challenges. We further introduce a cost-sensitive loss alleviating the dominance of non-semantic background candidates over the semantic candidates, thereby dealing with the data imbalance issue. Experimental results demonstrate that our tracker achieves competitive results at real-time speed compared to the state-of-the-art trackers.

\end{abstract}


\end{frontmatter}
\section{Introduction}\label{sec.1}
Visual object tracking aims to locate the target bounding box over a sequence of images after specifying the initial bounding box. Visual object tracking is a fundamental task for human-machine interaction, autonomous driving, visual sports analysis, virtual reality, and human motion analysis. These applications are mainly based on object position estimation. Visual tracking is a challenging vision task due to the limited information of the object of interest since only the first frame can be employed. The major challenges to be addressed include fast speed, motion blur, camera motion, occlusion, illumination change, appearance variation, and background clutter. To cope with these challenges, considerable works have gone into developing a powerful observation model based on the local structure of the object of interest. Convolutional Neural Networks (CNN) have recently gained popularity as a tool for learning reliable representations and complex models. They excelled at different vision tasks  such as object recognition, object identification, semantic segmentation, and so on. Owing to the powerful discriminative representation of deep CNNs, impressive progress has been made in CNN-based trackers; however, since the target suffers a wide range of unpredicted appearance variations over time, achieving high-accuracy tracking at real-time speed still remains an open problem. CNN-based visual trackers require a large amount of training data to achieve the generality of the expert models and robustness of the feature representation. However, very deep CNNs are less practical for one-shot learning tasks which aim to learn a model from a single input in an online mode. \par

\begin{figure}[!t]
 \centering
  \includegraphics[scale=0.33]{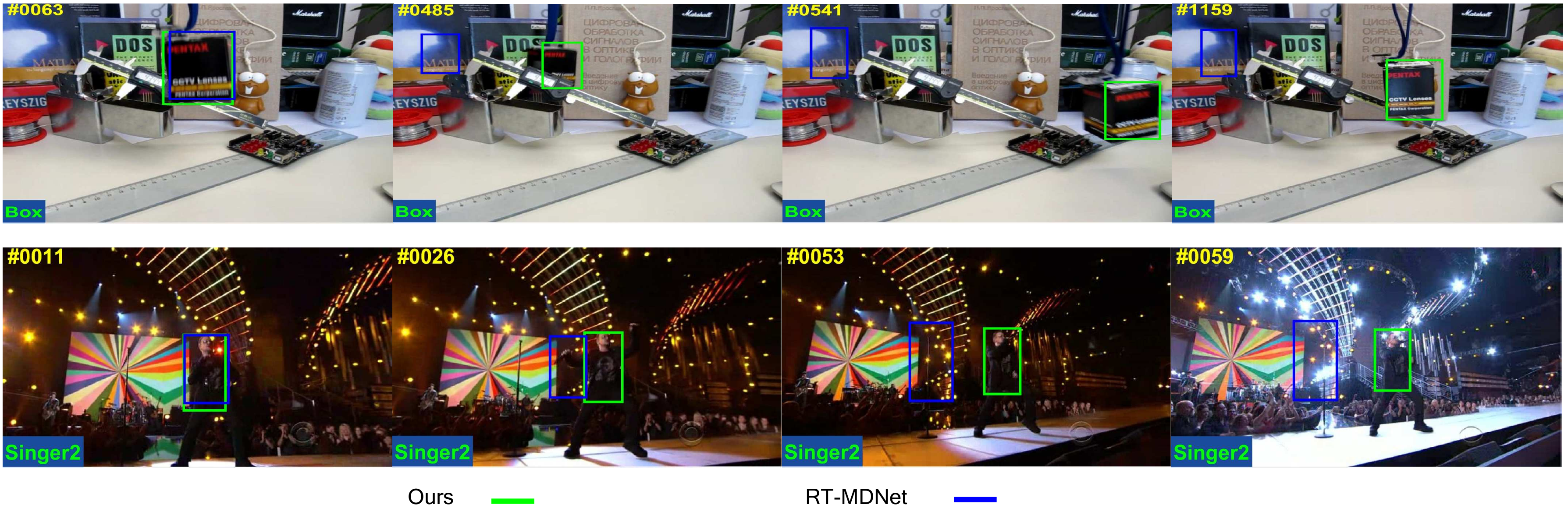}
  \caption{The comparison of our proposed tracker with RT-MDNet without a pre-training phase. The online learning policy in our tracker can more robustly deal with over-fitting issue in the challenging sequences in OTB dataset such as Singer2 and Box.}\label{fig.compare}
\end{figure}

Online learning policy has been demonstrated to be an effective approach in making visual trackers more robust against various distortions during tracking procedure \cite{danelljan2019atom,zhang2019learning,zhou2020discriminative}. Nevertheless, excessive online updating strategies could simply make the trackers prone to over-fitting to non-target context, resulting in tracking drift. Furthermore, naively integrating previous patch features into the long- and short-term feature templates discards the discriminative context of the background areas. To mitigate these issues, in this paper, we exploit the gradients of the positive and negative candidates through a cross-entropy loss function to capture the context-aware CNN filters for online updating. Armed with the selected convolutional filters, our tracker can accurately discriminate the semantic background candidates (also called distractors) from the positive ones at each time step. Additionally, reducing the number of parameters by such an efficient perspective may help to alleviate the over-fitting issue in updating phase. In contrast to other trackers \cite{nousi2020dense,kashiani2019visual,lu2020dense,gao2019graph} which conventionally have been pre-trained by means of large-scale object tracking datasets, we leverage only an off-the-shelf CNN model as feature extraction without offline pre-training procedure.  Thanks to the selection of context-aware CNN filters from the off-the-shelf CNN model, our tracker exhibits competitive results compared to the state-of-the-art visual trackers (as shown in Figure \ref{fig.compare}). Despite the fact that the off-the-shelf CNN models pre-trained for object classification tasks are agnostic of the intra-class discrepancies, the proposed selection approach can extract the context-aware CNN filters which contribute to the intra-class discrepancies.
As a result, the domain of the off-the-shelf CNN model is adapted for object tracking task with a different domain. In addition, as the off-the-shelf models do not require a pre-training phase, enjoying the advances in connection to the hand-engineered deep network architectures could be more feasible with our proposed strategy.\par

Despite the improvement made by the mentioned online learning policy for distinguishing an object of interest from foreground objects, drifting to distractors is not solved thoroughly in our real-time tracker. This mainly stems from the data imbalance issue, one of the deep-rooted issues in visual trackers \cite{fan2019siamese,zhu2018distractor,jung2018real,nam2016learning}. Data imbalance issue makes many trackers suffer poor generalization. In visual object tracking, data imbalance issue exists in two respects. First, the positive candidates form a significant proportion of training data in comparison with the negative ones, which adversely affects the performance of the CNN models. Second, the domination of non-semantic background candidates (i.e., the easy negative candidates) over the distractors makes the trained network to be biased toward easy non-semantic background candidates and degrades the performance of the network. Inspired by the recent advances in object detection and tracking \cite{lin2017focal,lu2018deep}, a cost-sensitive loss function is proposed to balance the contribution of non- and semantic background candidates and also positive/negative candidates in the updating strategy. That is to say, the proposed loss function manages to penalize the easy non-semantic background candidates while strengthening the impact of distracters, including the negative and positive ones. It is noteworthy that the proposed filter selection strategy also contributes to coping with the second aspect of the data imbalance issue in that it extracts the context-aware filters beneficial for discriminating the distractors from an object of interest.

We summarize our main contributions as follows:

\begin{itemize}
\item We propose a context-aware domain adaptation for online learning policy in our visual tracker so that the convolutional filters pre-trained in a different domain can be extracted for visual tracking, considering the context of different negative candidates. To this end, only the ground-truth of the first frame is utilized.
\item We propose a cost-sensitive loss function to ameliorate the fragility of our visual tracker against the data imbalance issue in the online learning procedure.

\item  We carry out extensive experimental evaluations to demonstrate that the online learning policy adopted in our tracker can achieve competitive performance without any pre-training phase in comparison to the state-of-the-art real-time trackers.
\end{itemize}

The rest of this paper is organized as follows: Section \ref{sec.2} gives a literature
review of the state-of-the-art object trackers based on deep learning and online learning policy. Section \ref{sec.3} offers a detailed description of our proposed approach for online learning policy and the cost-sensitive loss function, and Section \ref{sec.4} describes the baseline tracking algorithm. Section \ref{sec.5} presents extensive assessments on different datasets to
evaluate the contribution of the proposed components in our tracker, and Section \ref{sec.6} concludes our work
by some insightful points.

\section{Related Work}\label{sec.2}
In this section, the deep learning-based visual trackers and the trackers based on online learning policy are discussed in more detail.

\subsection{Deep Trackers}
In recent years, thanks to deep learning breakthroughs, visual object tracking \cite{valmadre2017end,li2020noise,wang2019unsupervised,bertinetto2016staple,li2014scale,danelljan2016discriminative} has enjoyed many advances the same as other areas in computer vision.
Generally speaking, the deep learning-based trackers are categorized into one-stage \cite{held2016learning,tao2016siamese,bertinetto2016fully,yan2019adaptive} (matching-based) and two-stage (classification-based) \cite{chen2018real,song2018vital,han2017branchout,nam2016learning} groups. One-stage trackers take advantage of a pre-trained CNN model to locate the most similar region of interest to the predefined template over time. Recently, transformer architectures have been gaining popularity in natural language processing. Transformer architecture employs attention-based encoders and decoders to transform a sequence into another one. The attention-based encoders and decoders exploit the global information from the input sequence by examining the input sequence and determine which parts of the sequence are relevant to another one.
Transformer architectures have increasingly expanded to deal with non-sequential problems after replacing recurrent neural networks in several sequential tasks such as natural language processing \cite{devlin2018bert}, voice processing \cite{synnaeve2019end}, and computer vision \cite{SETR,liu2021multimodal,sun2021loftr,Prakash2021CVPR}. In visual object tracking problem, authors in \cite{Transformer} exploit the transformer architectures to link the isolated frames in the video flow together and capture the rich temporal cues across frames.\par

In recent years, Siamese networks have also attracted much attention in the realm of one-stage trackers. Formulated as a cross-correlation problem, Siamese-based trackers \cite{fan2019siamese,li2019siamrpn++,guo2020siamcar,cao2019visual,yu2020deformable} train two-branch CNNs to encode search region and target patch simultaneously. In the inference stage, some Siamese-based trackers \cite{zhang2019learning,zhu2018distractor,choi2019deep,li2019gradnet} update their models to boost their robustness. The others \cite{held2016learning,li2018high,wang2018learning}, based on one-shot learning, fully discard updating phase for higher efficiency at lower accuracy cost. SiamFC \cite{bertinetto2016fully}, as the pioneering work in the Siamese-based trackers, pre-train a fully convolutional Siamese network to calculate a single-channel response map for object tracking without any updating phase. Similarly, \cite{tao2016siamese} and \cite{held2016learning} pre-train Siamese networks with different structures for online tracking. Following SiamFC \cite{valmadre2017end}, CFNet incorporates a correlation filter layer into the SiamFC model and updates its model by applying an average template. SiamRPN \cite{li2018high} integrates Region Proposal Network \cite{girshick2015fast} into the Siamese network, whereby classification  and regression branches can be trained jointly in the offline phase. DaSiamRPN \cite{zhu2018distractor} attempts to deal with the data imbalance issue in the SiamRPN and boosts its adaptability and also extends SiamRPN to the long-term tracking with local-to-global search region strategy. Authors in \cite{li2019siamrpn++} enhance the SiamRPN tracker to enjoy more abstract representation with deeper networks such as a modified version of ResNet-50 \cite{he2016deep}. Besides, the up-channel cross-correlation layer is supplanted by a depth-wise cross-correlation layer in their network to reduce the computational cost and yield better performance.\par

In comparison to the one-stage category, two-stage trackers discern the target from the background areas through pre-trained correlation filters or CNN-based classifiers. In the first stage, several candidates are drawn around the previous position of the target. These candidates are then evaluated and classified through a trained CNN model in the second stage.
Among two-stage trackers, MDNet \cite{nam2016learning} drastically yields huge performance gain in comparison with other trackers in 2016. MDNet integrates an online refining network into the tracking process to make the tracker more versatile in addressing different challenges, including appearance variations, background clutter, and occlusion. Based on its approach, numerous studies have been conducted so far \cite{jung2018real,song2018vital, han2017branchout,park2018meta,8740907,gao2020recursive,9035457}. Although MDNet has achieved top-ranked performance, several drawbacks have still remained to be addressed. The first one is associated with its high computational complexity. MDNet evaluates candidates independently through an offline pre-trained network and refines this network over time with sequence-specific information. The high computational cost of online refinement remarkably decreases the tracking speed, impeding its real-time application. Some studies have been conducted to mitigate this drawback. Authors in \cite{jung2018real} propose the state-of-the-art Real-time MDNet, called RT-MDNet, to speed up MDNet utilizing an adaptive RoI alignment (RoIalign) after the \textsl{conv3} layer to output a fix-sized shared feature map for all sampled candidates. Chen et al. \cite{chen2018real} formulate the tracking problem as an \textsl{Actor-Critic} framework, in which the actor model is pre-trained based on the reinforcement learning to predict one action at each time step during tracking. Such prediction is assessed by the Critic model in online and offline phases. MetaRTT \cite{jung2020real} tries to speed-up the model adaptation in online fashion employing one-shot network pruning with meta-learning.

\subsection{Deep Trackers with Online Learning}
Online updating phase makes visual trackers more adaptable in addressing object appearance variation, illumination changes, background clutter, and other challenging uncertainties. For this objective, various approaches have been employed so far, including incremental subspace \cite{ross2008incremental,wang2012online}, template integration \cite{zhu2018distractor,li2019gradnet,guo2017learning,zhu2018end,zhang2019learning}, gradient-based updating \cite{li2019gradnet, wang2015visual,song2017crest}, online classifier updating \cite{jung2018real,nam2016learning,danelljan2019atom,song2018vital}, and meta-learning optimization \cite{park2018meta,choi2019deep,dai2020high,jung2020real}. Most of the trackers in the template integration category adopt a fixed updating strategy with a linear interpolation policy.
With the limitations of such a naive strategy in mind, \cite{zhang2019learning} trains a two-layer CNN to learn how to update its templates non-linearly over time. The gradient-based updating category updates its model during the tracking process using gradient information. Li et al. \cite{li2019target} learn target-aware deep features for a Siamese network through the gradients captured with regression and ranking loss functions. Using discriminative information of the gradients in the feed-forward and backward operations, GradNet \cite{li2019gradnet} updates its template for a Siamese network. Concerning online classifier updating, the trackers train powerful classifiers in an online mode using the sampled candidates wrapping around the object of interest. ATOM \cite{danelljan2019atom} trains a 2-layer fully CNN to output a 2D-location of the target while using an optimization strategy, based on Conjugate Gradient and Gauss-Newton. In comparison to ATOM, trackers such as \cite{jung2018real,nam2016learning} with multi-domain pre-training phase update their models at each time step in online mode to classify foreground instances from the background ones with a cross-entropy loss function. Despite achieved high-accuracy performance, they pre-train their models through sophisticated multi-domain learning, thereby enjoying a wide range of learned information thanks to the available large-scale datasets (note that in our paper, we seek to gain the same competitive performance exploiting only the first frame without a pre-training phase). The last category, meta-learning based approach, generally pre-trains a meta-learner optimizer to speed-up convergence time \cite{jung2020real,choi2019deep,park2018meta}.\par

Even though an online updating strategy can make a tracker more robust in handling a wide range of variations in background and instance level, it can also cause drifting problems. To be more specific, inaccurate tracking predictions may simply introduce adverse noise into the updating phase, resulting in drifting issues. Moreover, due to the limited online training samples during tracking procedure, updating strategy could exacerbate the fragility of the trackers against the over-fitting issue. In short, it can be a double-edged sword for visual tracking problem. In this paper, we launch a study to explore to what extent we can leverage online learning policy (in the online classifier updating category) without enduring the mentioned challenges. Using the gradient computed of the negative candidates, our online updating policy deal with the over-fitting problem and emphasize on more discriminative background candidates in the online learning process.

\section{The Proposed Tracker}\label{sec.3}
In this section, first, the architecture of our model is presented. Then, we explain how an off-the-shelf CNN can be well adapted to our visual tracker without any pre-training phase. Finally, a cost-sensitive loss function is also introduced to alleviate the data imbalance issue in the online learning policy.
\begin{figure}[htb]
\captionsetup[subfigure]{justification=centering}
    \centering
    \begin{subfigure}[b]{\textwidth}
        \centering
        \includegraphics[width=1.1\linewidth]{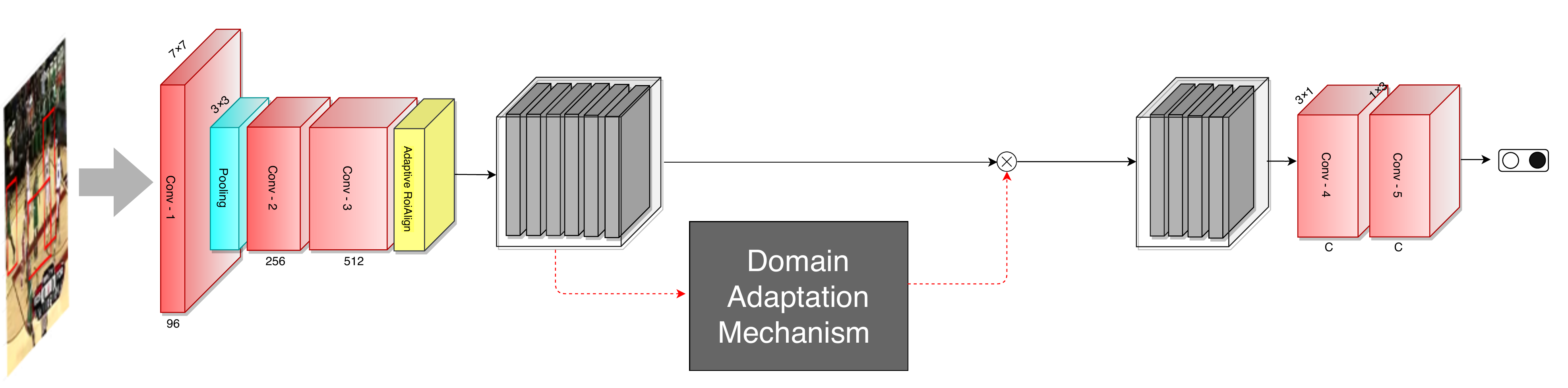}%
        \vspace{.5mm}
        \caption{}
    \end{subfigure}
    \vskip\baselineskip
    \begin{subfigure}[b]{\textwidth}
        \centering
        \includegraphics[width=0.9\linewidth]{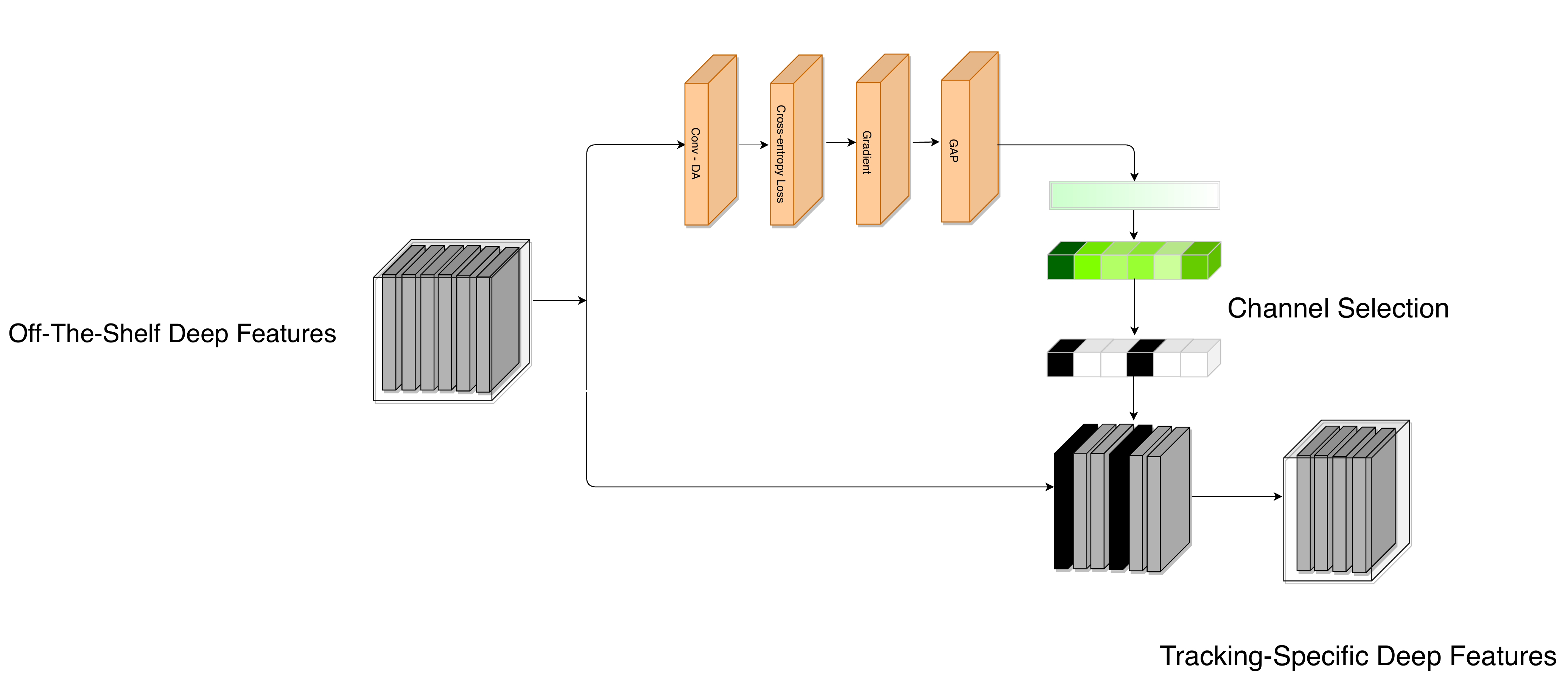}%
        \caption{}
    \end{subfigure}
  \caption{The proposed architecture of our network with domain adaptation. (a) Network Architecture. (b) Domain Adaptation Mechanism. Best viewed in color and magnification.}
\label{fig.architecture}
\end{figure}
\subsection{Network Architecture}
As depicted in Figure \ref{fig.architecture} (a), our network is a fully convolutional neural network, the feature extractor of which is backboned by VGG-M \cite{chatfield2014return}. More specifically, the feature extractor consists of three convolutional layers (i.e., Conv-1 to Conv-3), one max-pooling layer, and one adaptive RoIAlign layer. All convolutional layers are followed by ReLU and local response normalization (LRN). The second max-pooling layer in the VGG-M network is eliminated, and the dilation rate of Conv-3 is set to 3. The adaptive RoIAlign layer is employed after Conv-3 to compute feature of each RoI sampled during tracking same as RT-MDNet \cite{jung2018real}. For online learning, we do not adopt the domain-independent subnetwork of RT-MDNet in that three fully connected (FC) layers with a large number of parameters can make our tracker susceptible to over-fitting to non-semantic background areas. We replace fully connected (FC) layers with two stacked $3\times1$ and $1\times3$ convolutional layers (called Conv-4 and Conv-5), as shonw in Figure \ref{fig.architecture} (a). Ultimately, to decrease the channel number to the number of classes, a $1\times1$ convolutional layer (Conv-6) is also applied to the network.

\subsection{One-Shot Domain Adaptation}
GradNet \cite{li2019gradnet} proves that the absolute value of the gradients is higher for the distractors pixels than the non-semantic background areas. Inspired by GradNet and relevant studies \cite{zhou2016learning,li2019target}, we train a one-layer $3\times3$ convolution layer, named Conv-DA (Figure \ref{fig.architecture} (b)), with a cross-entropy loss function in the first frame to emphasize the semantic background candidates for channel selection. The selected channels retain more discriminative background information advantageous for online learning policy. Furthermore, such channels can ameliorate the fragility of online learning against the over-fitting to the recent non-semantic background areas. To this end, in the first frame of a video sequence, different candidates are sampled and fed to the network, as illustrated in Figure \ref{fig.architecture} (a). Once the features of all sampled candidates are calculated in the first frame with the adaptive RoIAlign, the Conv-DA is trained with a cross-entropy loss function. Then, the gradients of
the scores for the background class are calculated regarding the feature map activations according to Figure \ref{fig.architecture} (b). Finally, based on \cite{zhou2016learning}, we can select the appropriate channels to capture the context-aware feature space for domain adaptation (from classification to tracking) by employing the global average pooling over the width and height dimensions. The global average pooling operation for the gradients of the scores in the background class is calculated as follows:

\begin{equation}\label{eq.0}
  {\delta}^{n}_{k}=\frac{1}{N}\sum_{i=1} \sum_{j=1} \: \frac{\partial L}{\partial C_{k}},
\end{equation}

\noindent  where $\delta^{n}_{k}$ denotes the importance of n-th channel, $ L$  is the cross-entropy loss function, $N$ denotes the number of feature elements in $C_{k}$, and $C_{k}$ indicates the feature of n-th channel fed to the one-later convolution layer in the first frame.  Figure \ref{fig.compare} illustrates the performance of online learning in RT-MDNet tracker without a pre-training phase. As shown in Figure \ref{fig.compare}, owing to the large number of parameters needed to be adjusted by limited sampled candidates, RT-MDNet fails to track an object of interest without pre-training phase and over-fits to unfavorable areas. The failures mostly occur when there are less discriminative background areas compared to the target regions. However, the domain adaptation policy can handle such a tricky task through capturing the context-aware CNN filters so that the inter- and intra-class discrepancies would be tailored for discriminating background areas from the target.

\subsection{Cost-Sensitive Loss}
One of the key remaining demerits of two-stage trackers is concerned with their incapabilities to cope with the data imbalance issue. In this paper, to deal with this issue, a cross-entropy (CE) loss is reformulated to eliminate the class biases in the online learning process. In this regard, the CE loss is initially formulated as \cite{lin2017focal}:
\begin{equation}\label{eq.1}
CE(p,y)=-\log(p_{t}),
\end{equation}

\begin{equation}\label{eq.2}
p_{t}=\begin{cases}
    \;p,& \text{if $y=1$}\\
    \;1-p,& \text{otherwise},
\end{cases}
\end{equation}
\noindent where $p\in [0,1]$ denotes the probability of each candidate, and $y \in\{0,1\}$ indicates ground-truth labels. The multiplicity of easy background candidates ($p_{t} \ll 0.5, y = 0$) over rare foreground candidates ($y = 1$) and also over hard background candidates ($p_{t} \gg 0.5, y = 0$) makes the losses generated by the CE prone to easy background candidates. In dealing with such issue, inspired by \cite{lin2017focal}, VITAL \cite{song2018vital} incorporates a probability-dependent term into the CE loss as below:\\
\begin{equation}\label{eq.3}
L(p_{t}) = -(1-p_{t})^{\nu}\log(p_{t}),
\end{equation}
\noindent where $\nu> 0$ indicates a tunable focusing parameter, which is set to 1. Equipped with this loss function, VITAL precludes easy background candidates from dominating the gradient. However, it down-weights the losses of the hard background candidates in addition to the easy background ones; thus, it fails to exploit the precious information of distracters and merely gain a slight improvement.
To penalize easy non-semantic background candidates and take into account the contributions of the distracters, we integrate a new modulating term into the CE loss as below:
\begin{equation}\label{eq.4}
L(p_{t}) = \frac{-\log(p_{t})}{(1+\exp (\alpha(\beta-(1-p_{t})^{\gamma})))},
\end{equation}
\noindent where $\alpha$, $\gamma$, and $\beta$ are hyper-parameters, regulating the amount and the location of candidates penalization. The proposed modulating term down-weights easy candidates ($p_{t} \gg 0.5$) while trying to keep hard candidates ($p_{t} \ll 0.5$) unchanged.  As $pt\rightarrow1$, the modulating term decreases and as $pt\rightarrow0$  the modulating term keeping the impact of hard candidates unchanged. \par

With the proposed loss function, our tracker can be trained in the online fashion with the Stochastic Gradient Descent (SGD); as a result, it can alleviate the dominance of easy negative candidates over hard negative and positive ones, resulting in coping with the over-fitting issue. With such a hand-engineered loss and domain adaptation strategy, sequence-specific context can be elaborately modeled in the Conv-5 to Conv-6 layers. Therefore, our tracker manages to take into consideration target appearance variations, background clutter, and distracter objects without enduring over-fitting issue.

\section{Tracking Algorithm}\label{sec.4}
In this section, first, our tracker algorithm and sampling scheme are described. Then, the long- and short-term strategies for updating phase in our tracker are explained in detail.

\subsection{Online Tracking Procedure}
In the initial frame, for domain adaptation, positive and negative candidates are sampled to be fed to the Conv-DA layer with the cross-entropy loss function. The best channel indices for object tracking task are determined through Equation \ref{eq.0}. Then, Conv-5 to Conv-6 layers are fine-tuned as well. It is worth highlighting that for domain adaptation, negative candidates are sampled with different radius compared to the negative candidates adopted for the fine-tuning procedure. This is ascribed to the fact that the negative candidates in the domain adaptation phase should contain different background regions from which the object may pass in the following frames. From the second frame on, Conv-4 to Conv-6 layers are fine-tuned under a peculiar condition for capturing target appearance variations. Like other two-stage trackers, at each frame, several candidates are sampled from a Normal distribution at the center of the previous target state. These candidates are then fed into the network, and the candidate with the highest classification score can be determined as the target as follows:

\begin{equation}\label{eq.5}
 \
x^\ast _{t}=  \argminA_{{x_{t}^i}} f^{+}(x_{t}^i),
\end{equation}

\noindent where $f^{+}(x_i)$ denotes the positive score of $i$-th sampled candidate at time step $t$. Employing the extracted features for different ROIs from the RoiAlign layer, a simple bounding box regressor is trained using 1000 candidates in the initial frame. This regressor is only utilized in reliable conditions.

\begin{spacing}{1}
 \linespread{1.25}
\begin{algorithm}[!t]
  \begin{algorithmic}[1]
    \Input{ Pretrained VGG-M, The first ground truth bounding box $x_g$.}
    \Output{Targat location $x^\ast _{t}$
    \newline Fine-tune: Conv-4, Conv-5, Conv-6.
    \newline Trained: Conv-DA, Conv-4, Conv-5, Conv-6.}
    \State Randomly initialize the weights of Conv-DA, Conv-4, Conv-5, Conv-6.
   \State Generate candidates (${m}_{1}^{+}$, ${m}_{1}^{-}$) around the first target position $x_g$.
   \State  \multiline{Feed ${m}_{1}^{+}$, ${m}_{1}^{-}$ to the off-the-shelf subnetwork, i.e. Conv-1, Conv-2, Conv-3.}
    \State \multiline{Train Conv-DA using the off-the-shelf deep features.}
    \State Calculate the tracking-specific deep feature using Equation(\ref{eq.0}).
    \State Train Conv-4, Conv-5, Conv-6 using the tracking-specific deep features.

    \State \multiline{  $\mathcal{M}_{short}\leftarrow{1}$ , $\mathcal{M}_{long}\leftarrow{1}$ and $x^\ast _{1}\leftarrow{x_g}$.}
    \For {$t = 2,3,...$  }
        \State \multiline{Extract sample candidates around $x^\ast _{t}$.}
    \State Calculate the tracking-specific deep feature using Equation(\ref{eq.0}).
        \State \multiline {Calculate $x^\ast _{t}$ using Equation(\ref{eq.5}).}
        \If {$ f^{+}(x^\ast _{t}) > 0.5$}
            \State \multiline{Generate candidates around the $x^\ast _{t}$ and pass them through the network.}
            \State $\mathcal{M}_{short}\leftarrow\mathcal{M}_{short}\cup\{t\}$, $\mathcal{M}_{long}\leftarrow\mathcal{M}_{long}\cup\{t\}.$

            \If {$|\mathcal{M}_{long}|>{\tau}_{long}$} {$\mathcal{M}_{long}\leftarrow\mathcal{M}_{long}\backslash\{ \min\nolimits_{\forall u \in {\mathcal{M}_{long}}} u \}.$}
             \EndIf

            \If {$|\mathcal{M}_{short}|>{\tau}_{short}$} {$\mathcal{M}_{short}\leftarrow\mathcal{M}_{short}\backslash\{ \min\nolimits_{\forall u \in {\mathcal{M}_{short}}} u \}.$}
            \EndIf
        \EndIf
        \If {$ f^{+}(x^\ast _{t}) < 0.5$}
            \State \multiline{Fine-tune Conv-4, Conv-5, Conv-6 using ${m}_{t}^{+} {_{\forall u \in {\mathcal{M}_{short}}}}$ and ${m}_{t}^{-} {_{\forall u \in {\mathcal{M}_{short}}}}$}
        \ElsIf {$mod(t,{\tau}_{int})=0$}
            \State \multiline{Fine-tune Conv-4, Conv-5, Conv-6 using ${m}_{t}^{+} {_{\forall u \in {\mathcal{M}_{long}}}}$ and ${m}_{t}^{-} {_{\forall u \in {\mathcal{M}_{short}}}}$}
        \EndIf
    \EndFor
  \end{algorithmic}
  \caption{Tracking Algorithm}
  \label{tracking_alg}
\end{algorithm}
\end{spacing}

\subsection{Updating Strategy}
In real-world scenarios, a target usually undergoes various challenges, including appearance variations, object deformation, illumination changes. A robust visual tracker should be fine-tuned during tracking procedure to take into account all these challenges. In this regard, to make our tracker robust and adaptive in dealing with the mentioned challenges, we equip our tracker with short- and long-term updating policies as MDNet. In the long-term strategy, updating is performed every ${\tau}_{int}$ frames with the positive candidates gathered from the previous successful frames in the frame set $\mathcal{M}_{long}$. For the short-term strategy, updating is executed as long as $ f^{+}(x_{t}^i)$ does not reach a predetermined threshold. The positive candidates in the short-term updating are gathered from the previous ${\tau}_{short}$ successful frames in the frame set $\mathcal{M}_{short}$. It is worth noting that in both strategies, the negative candidates are gathered from the previous ${\tau}_{short}$ successful frames. The whole procedure of our proposed online tracking is presented in \mbox{Algorithm \ref{tracking_alg}}.\par

\section{Experimental Results}\label{sec.5}
In this section, first, the experimental settings and implementation details are explained. Then, we carry out quantitative and qualitative experiments to evaluate our proposed tracker in comparison with the state-of-the-art visual trackers. To do so, our tracker is evaluated on the popular visual tracking datasets, namely OTB-2013 \cite{wu2013online}, OTB-2015 \cite{wu2015object}, OTB-50 \cite{wu2015object}. All the experimental results are conducted on a single NVIDIA Geforce GTX 1080 TI GPU with 11GB memory and PyTorch toolbox. The average tracking speed for our proposed method is approximately 24 FPS.

\subsection{Implementation details}
\paragraph{Tracker Settings} For online training, the first three layers weights are totally transferred from VGG-M, and all other layers, including Conv-4,  Conv-5, and  Conv-6 are initialized randomly. The target size is set to $107\times107$, and the input image is resized with the same scale value. In the initial frame of the tracking process, 5000 negative (${m}_{1}^{-}$) and 500 positive (${m}_{1}^{+}$) candidates are sampled to train the convolutional layers for 50 iterations with a learning rate of 0.0015. For the following frames, 200 negative (${m}_{t}^{-}$) and 50 positive (${m}_{t}^{+}$) candidates are collected from successfully tracked frames to fine-tune the convolutional layers for 10 iterations with a learning rate of 0.0025. In all frames, each mini-batch includes 32 positives and 96 negatives examples. In the first frame, sampled candidates are deemed to be positive when their IoU overlap ratios with the ground truth exceed 0.7, and also they are supposed to be negative if their IoU overlap ratios do not exceed 0.5. From the second frame on, the IoU overlap ratio for the negative candidates is changed to 0.3.

\paragraph{Domain Adaptation Settings}For domain adaptation, the learning rate, the maximum iteration number, and the number of candidates are set to 0.003, 100, and 500. We select the top 420 important channels through Equation \ref{eq.0} for learning semantic context. The weights of the one-layer convolution layer in the domain adaptation phase are initialized randomly.

\subsection{Evaluation on OTB}
\subsubsection{Dataset and Evaluation Metrics}
OTB \cite{wu2015object,wu2013online} is one of the popular tracking benchmarks with 11 different challenges, containing motion blur, appearance variation, occlusion, deformation, fast motion, background clutter. In this paper, we make evaluations in different editions of OTB benchmark, including OTB-2013, OTB-50, and OTB-2015. OTB-2013, OTB-50, and OTB-2015 are composed of 51, 50, and 100 different video sequences, respectively. The video sequences in the OTB-50 dataset consists of more challenging data compared to the OTB-2013 dataset. To assess our proposed approach in comparison with the state-of-the-art trackers, we adopt the evaluation plots proposed in \cite{wu2015object,wu2013online}, namely precision and success plots. These plots are drawn by means of distance precision (DP) and overlap success (OS) criteria. The precision plot illustrates the ratio of the frames whose center location error is within 20 pixels. The success plot calculates the percentage of the successfully tracked frames whose overlap criteria is larger than a predetermined threshold. The area under the curve (AUC) of this plot is taken into account for ranking purposes. In our evaluations, the one-pass evaluation (OPE) in OTB toolbox is employed to assess our algorithm compared to the state-of-the-art works, including CREST \cite{song2017crest}, SiamFC \cite{bertinetto2016fully}, CFNet \cite{valmadre2017end},  HCFTs \cite{ma2018robust}, TRACA \cite{choi2018context}, ACFN \cite{choi2017attentional}, SRDCF \cite{danelljan2015learning}, and Staple \cite{bertinetto2016staple}, BranchOut \cite{han2017branchout},  DSLT \cite{lu2018deep}, P2P \cite{gao2018p2t}, LCT \cite{ma2018adaptive}, AdaDDCF \cite{han2018adaptive}, Corrective \cite{8740907}, Quad \cite{dong2019quadruplet}.

\subsubsection{Internal Comparison}
\begin{figure}[!t]
  \centering
  \includegraphics[scale=1]{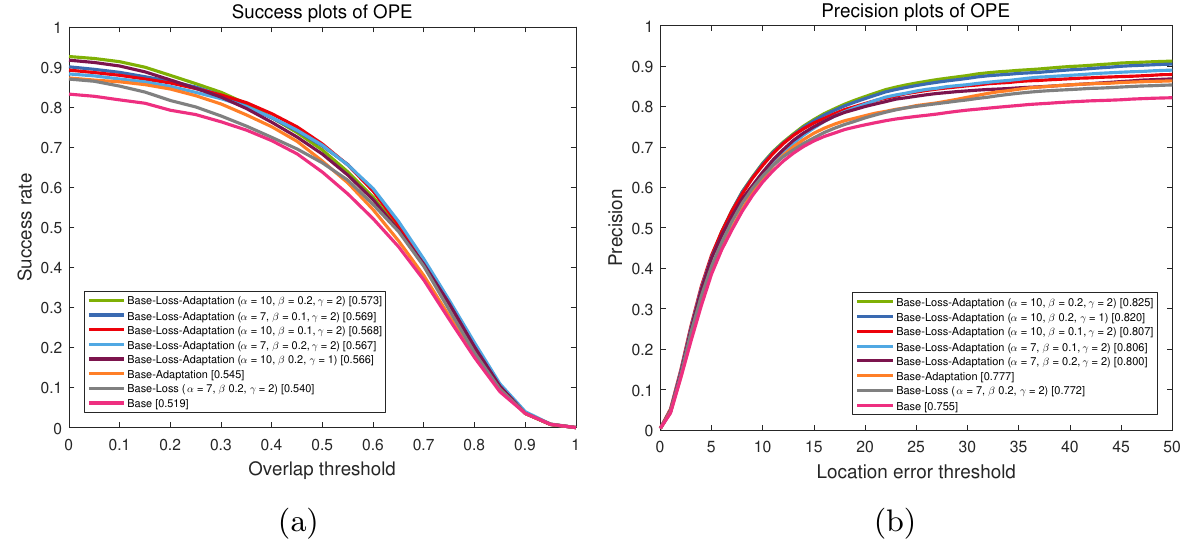}
  \caption{Results of self-comparison evaluation on the OTB-50 dataset. (a) Success plot, (b) Precision plot.}\label{fig.self_comparison}
\end{figure}

To investigate the effectiveness of domain adaptation and cost-sensitive loss components in our tracker, we deactivate them separately and assess the baseline of our tracker without them. In addition, to achieve the best performance, we tune different parameters in Equation \ref{eq.4}. As depicted in Figure \ref{fig.self_comparison}, both components boost the performance of our tracker against the baseline version, and apparently domain adaptation and cost-sensitive loss contribute equally to the tracker accuracy.  Figure \ref{fig.self_comparison} demonstrates that the parameter setting $\alpha = 10 , \beta = 0.2, \gamma = 2$ yields the best performance in terms of accuracy criterion among its variants.

\subsection{Quantitative Comparison}
\begin{figure}[!t]
  \centering
  \includegraphics[scale=1.05]{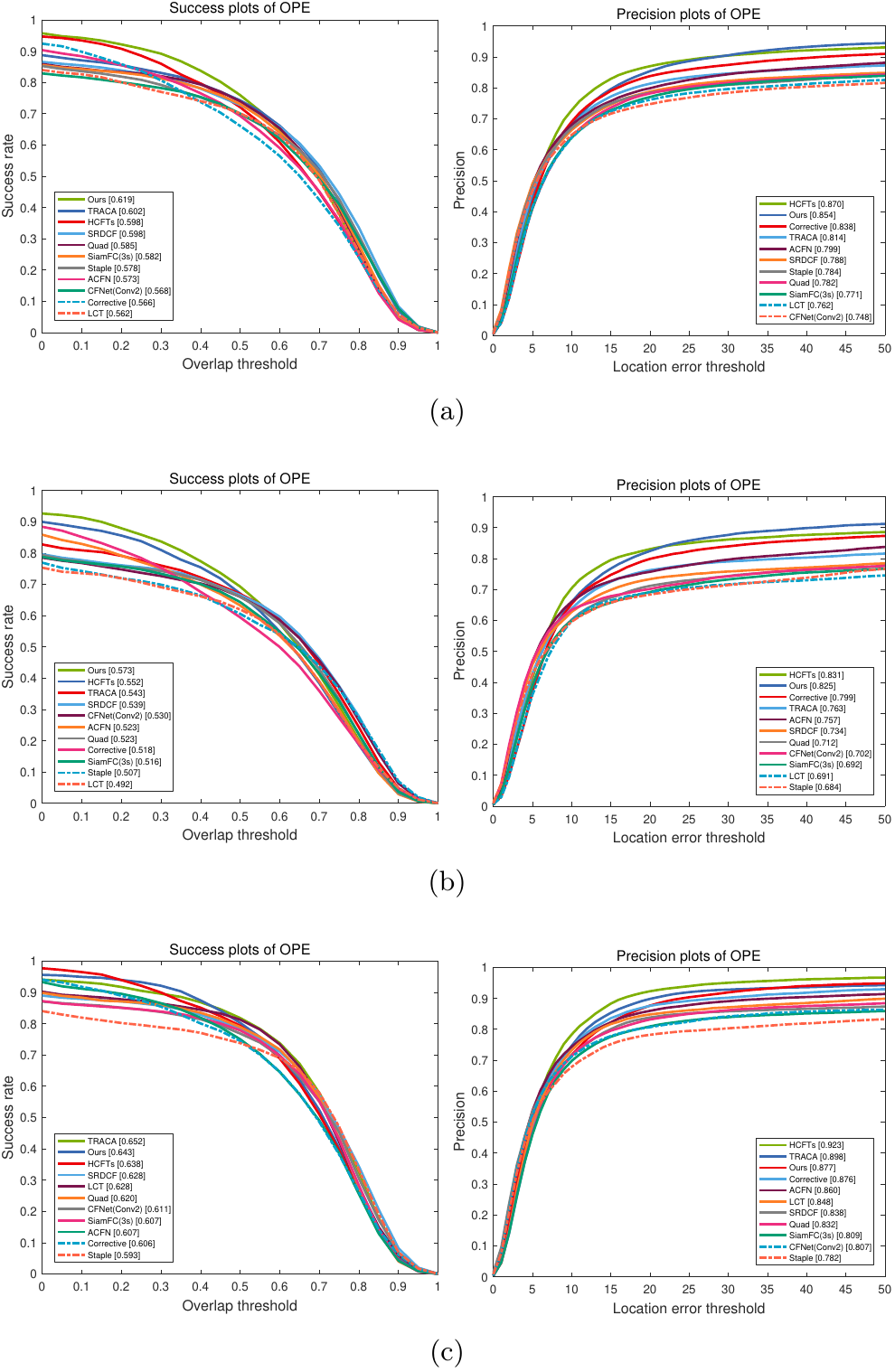}
     \caption{Quantitative results of our proposed tracker compared to its competitors in OPE evaluation on OTB benchmark. In the legend, the DP rates at the 20 pixels ratio and the AUC scores are reported. From top to bottom, we show the results on (a) OTB100, (b) OTB50, and (c) OTB2013 datasets.}\label{Fig.overalOTB}
\end{figure}

Since the scope of this work is limited to the visual trackers which are backboned with the off-the-shelf CNN models and are not specifically pre-trained for tracking task, first, we select the relevant state-of-the-art trackers to
make apples-to-apples comparisons. Due to the fact that the online learning phase is a time-consuming process for such trackers, most of these trackers are non real-time. Thus, we classify them into real-time and non-real-time trackers. Table \ref{table.1} reports the overall results of our tracker in comparison to its competitors on OTB100, OTB50, OTB2013 datasets. As shown in Table \ref{table.1}, our proposed tracker obtains satisfactory results compared to its competitors. To be more specific, in real-time class, our approach yields gains of 1.6\%, and 5.3\% on AUC and DP scores compared to the Corrective \cite{8740907} on OTB100 dataset. Despite the fact that BranchOut \cite{han2017branchout}, DSLT \cite{lu2018deep}, CREST \cite{song2017crest}, and P2P \cite{gao2018p2t} achieves better results in terms of OS and DP criteria, they fail to track an object of interest in real-time speed. That is, sophisticated, time-consuming online learning phases adopted in such trackers hinder their real-time applications. However, our tracker manages to achieve competitive performance while maintaining a real-time speed (24 FPS) thanks to its components, namely the domain adaptation and cost-sensitive loss. Excluding the mentioned non-real-time trackers, our proposed tracker is comparable to other trackers such as AdaDDCF \cite{han2018adaptive} and HCFTs \cite{ma2017robust} in terms of the OS and DP metrics while running at real-time speed. In addition to the comparison reported in  Table \ref{table.1}, to investigate the performance of our tracker over the state-of-the-art trackers pre-trained in offline mode, we also make another comparison. The trackers in this assessment include Quad \cite{dong2019quadruplet}, CFNet \cite{valmadre2017end},  TRACA \cite{choi2018context}, ACFN \cite{choi2017attentional}, to name but a few. Figure \ref{Fig.overalOTB} demonstrates that, for the most part, the superiority of our tracker is maintained even in comparison to the trackers leveraging from a wide range of information in offline training fashion. Besides, it is observed that our tracker performs better on the OTB-50 dataset over other trackers even though the OTB-50 dataset is more challenging than the OTB-2013 dataset.

\begin{table}[!t]
\centering
\tabcolsep 3pt
\caption[Caption for Quantitative results]{Comparison of online learning (OL) based trackers on the OTB100, OTB50 and OTB2013 datasets. Tackers are categorized into real-time (OL-RT) and non-real-time (OL-NRT) groups. The overlap success (OS) is calculated according to the AUC score, and the distance precision (DP) is reported at the error threshold of 20 pixels, respectively. The FPS results are reported based on the reference papers.}
\medskip\small
\resizebox{0.7\textwidth}{!}{
\begin{tabular}{LLccccccc}
\hline
  \hline
  \addlinespace[1mm]
   && \multicolumn{2}{c}{{OTB100}} & \multicolumn{2}{c}{{OTB50}} & \multicolumn{2}{c}{{OTB2013}}\\[1mm]
  \textbf&{Method} & {DP}  & {OS} & {DP} & {OS} & {DP}& {OS}& {FPS}\\[2mm]
\hline
  \addlinespace[2mm]
\multirow{2}{*}{\rotatebox[origin=c]{90}{\makecell{OL-RT}}}
&Ours                      & 85.4               & 61.9   &  82.5 & 57.3    & 87.7   & 64.3  & 24 \\ [1mm]
&Corrective \cite{8740907} & $83.8$             & $56.6$            & $79.9$ & $51.8$  & $87.6$ & $60.6$& $35 $\\ [1mm]

  \addlinespace[1mm]
\hline
  \addlinespace[1mm]
  \multirow{7}{*}{\rotatebox[origin=c]{90}{\makecell{OL-NRT}}}
&BranchOut \cite{han2017branchout}  & $91.7$  & $67.8$ & - & - & - & -& $1$\\[1mm]
&DSLT \cite{lu2018deep} & $90.9$ & $66$ & 87.4 & 62.1 & $93.4$ & $68.$ & $5$\\ [1mm]
&CREST \cite{song2017crest}& $83.8$  & $62.3$ & $79.1$ & $56.8$ & $90.8$ & $67.3$& $1$\\ [1mm]
&P2P \cite{gao2018p2t}& $85.4$  & $62.8$ & - & - & $90.8$ & $66.3$& $2$\\ [1mm]
&AdaDDCF \cite{han2018adaptive}& $87.2 $  & $61.2$ & $83.9$ & $57.7$ & $88.2 $ & $64.3$& $9$\\ [1mm]
&HCFTs \cite{ma2017robust} & $87$    & $59.8$   &  $83.1$ & $55.2 $ & $92.3$ & $63.8$& $6.7 $\\ [1mm]

  \hline
  \hline
  \label{table.1}
\end{tabular}}
\end{table}

\subsection{Attribute-Based Comparison}
\begin{figure}[!t]
  \centering
  \includegraphics[scale=1.05]{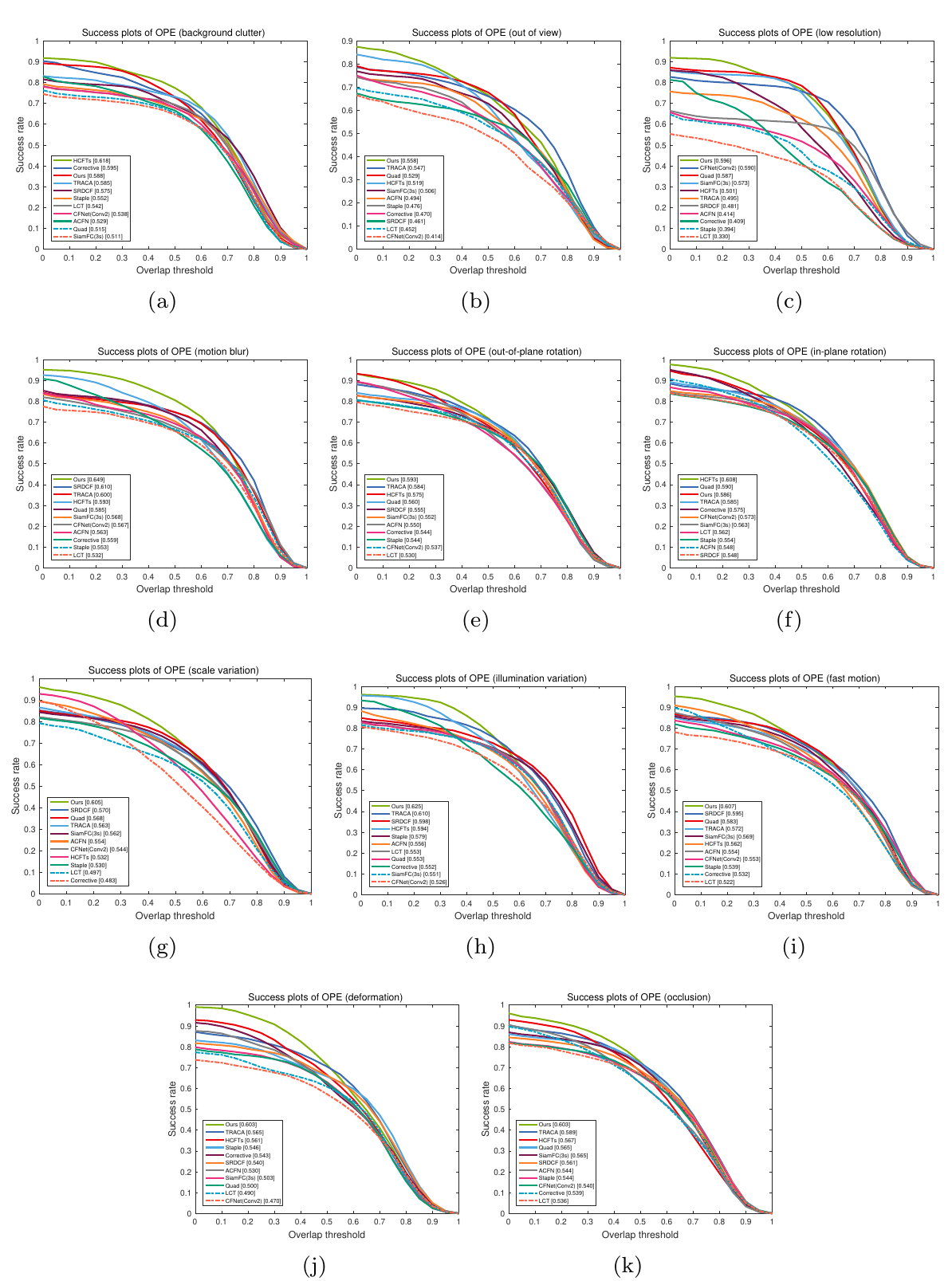}
   \caption{Success plots of OPE on 11 challenges, including (a) background clutter, (b) out-of-view, (c) low-resolution, (d) motion blur, (e) out-of-plane-rotation, (f) in-plane-rotation, (g) scale variation, (h) illumination variation, (i) fast motion, (j) deformation, and (k) occlusion.}\label{fig.AttributeReslts_succ}
\end{figure}

To investigate the robustness of our tracker against different challenges, we employ the per-attribute based evaluation on the OTB benchmark in which all video sequences are labeled and categorized with 11 different types of challenges. The challenges include fast motion, deformation, illumination variation, background clutter, out-of-plane rotation, low resolution, occlusion, scale variation. Figure \ref{fig.AttributeReslts_succ} compares the results achieved from our tracker over other studies with respect to the OS criterion on the OTB-100 dataset. Figure \ref{fig.AttributeReslts_succ} demonstrates that our tracker ranks first on 9 out of 11 per-attribute based evaluations. More specifically, the highest gains are related to the scale variation, deformation, motion blur, and illumination variation challenges. This is attributed to the domain adaptation components, which emphasizes the discriminative areas in background samples. However, such superiority is not maintained in background clutter against HCFTs \cite{ma2017robust} and Corrective \cite{8740907}, as illustrated in Figure \ref{fig.AttributeReslts_succ} (a). This is due to the fact that both HCFTs \cite{ma2017robust} and Corrective \cite{8740907} exploit deep feature hierarchies beneficial for extracting semantics and spatial context, thereby making them best in handling background clutter challenge. In relation to other challenges, our tracker also ranks first in out-of-view and occlusion categories with a modest gain over the runner-ups (see Figure \ref{fig.AttributeReslts_succ} (b) and \ref{fig.AttributeReslts_succ} (k)). Taking these results into account, we can conclude that our tracker possesses high robustness in coping with a wide range of challenges while running at real-time speed.

\begin{figure}[!t]
  \centering
  \includegraphics[scale=1]{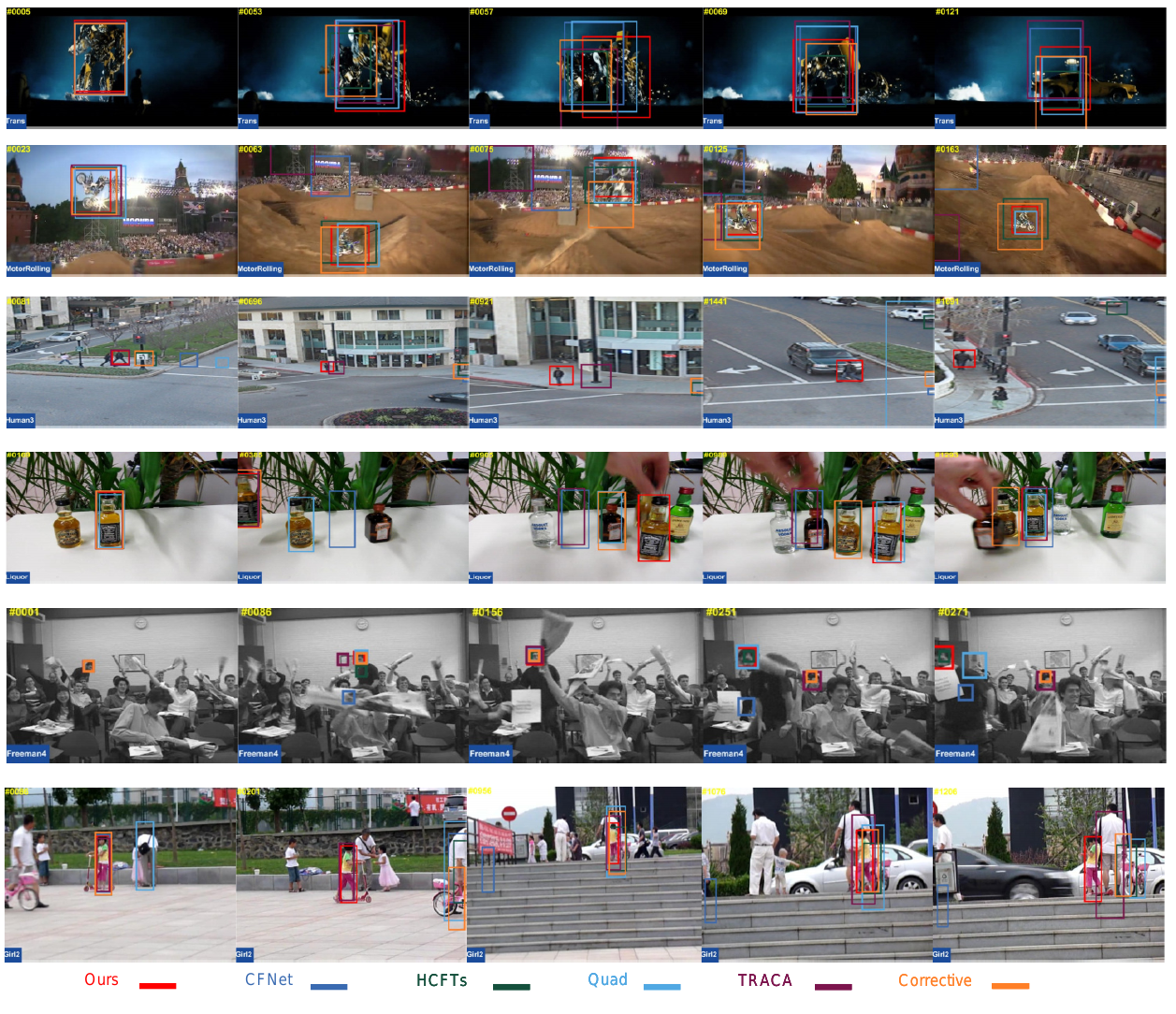}
  \caption{Qualitative performance of our proposed tracker on some challenging videos in the OTB100 dataset in comparison to the state-of-the-art trackers, including CFNet \cite{valmadre2017end}, HCFTs \cite{ma2018robust}, Quad \cite{dong2019quadruplet}, TRACA \cite{choi2018context}, and Corrective \cite{8740907} studies. (from top to bottom:  \textit{Trans}, \textit{MotorRolling}, \textit{Human3}, \textit{Liquor}, \textit{Freeman3}, \textit{Girl2})}\label{fig.BoundingBox}
\end{figure}

\subsection{Qualitative Comparison}
For qualitative evaluation, as represented in Figure \ref{fig.BoundingBox}, we opt for some challenging sequences in the OTB dataset with various difficulties, including deformation (\textit{Trans}, \textit{Human3}, \textit{Girl2}), scale variations (\textit{Trans}, \textit{MotorRolling}, \textit{Human3}, \textit{Liquor}, \textit{Freeman3}, \textit{Girl2}), occlusion (\textit{Human3}, \textit{Liquor}, \textit{Freeman3}, \textit{Girl2}), and background clutter (\textit{MotorRolling}, \textit{Human3}, \textit{Liquor}). In our assessments, CFNet \cite{valmadre2017end}, HCFTs \cite{ma2018robust}, Quad \cite{dong2019quadruplet}, TRACA \cite{choi2018context}, and Corrective \cite{8740907} trackers are selected to be compared with our tracker.
In the \textit{Trans} sequence, it is observed that all the trackers fail to fit the scale variations and deformation accurately. Even though CFNet, TRACA, and Quad have been pre-trained using large-scale datasets, they cannot handle the deformation issue. In the \textit{MotorRolling} sequence, CFNet, and TRACA fail to deal with severe background clutter and drift away. Nevertheless, our tracker and Quad robustly track the target and fit the scale variations more accurately. Confusing similar objects in the \textit{Human3} sequence substantially makes all the tracker drift away, while our method tackles such a tricky issue by virtue of the cost-sensitive loss component. The object of interest in the \textit{Liquor} sequence is frequently occluded by the other objects, whereby CFNet, HCFTs, TRACA, and Corrective trackers mistakenly locate the distractors. Owing to the cost-sensitive loss component, our approach manages to deal with confusing objects and estimate target location. The robustness of our tracker against occlusion and distractors is also retained in the  \textit{Girl2} sequence. Note that the channel selection in the domain adaptation strategy does not considerably contribute to distinguishing all kinds of new, unseen distractors from the target in that it is only applied in the first frame. As a result, the impact of domain adaptation is more pronounced for the \textit{Trans}, \textit{Liquor}, and \textit{Freeman3} sequences in which the background context is more stationary. In general, our tracker is able to robustly locate the targets in even complicated scenarios with a combination of different challenges.

\subsection{Failure Cases}
\begin{figure}[!t]
 \centering
  \includegraphics[scale=0.7]{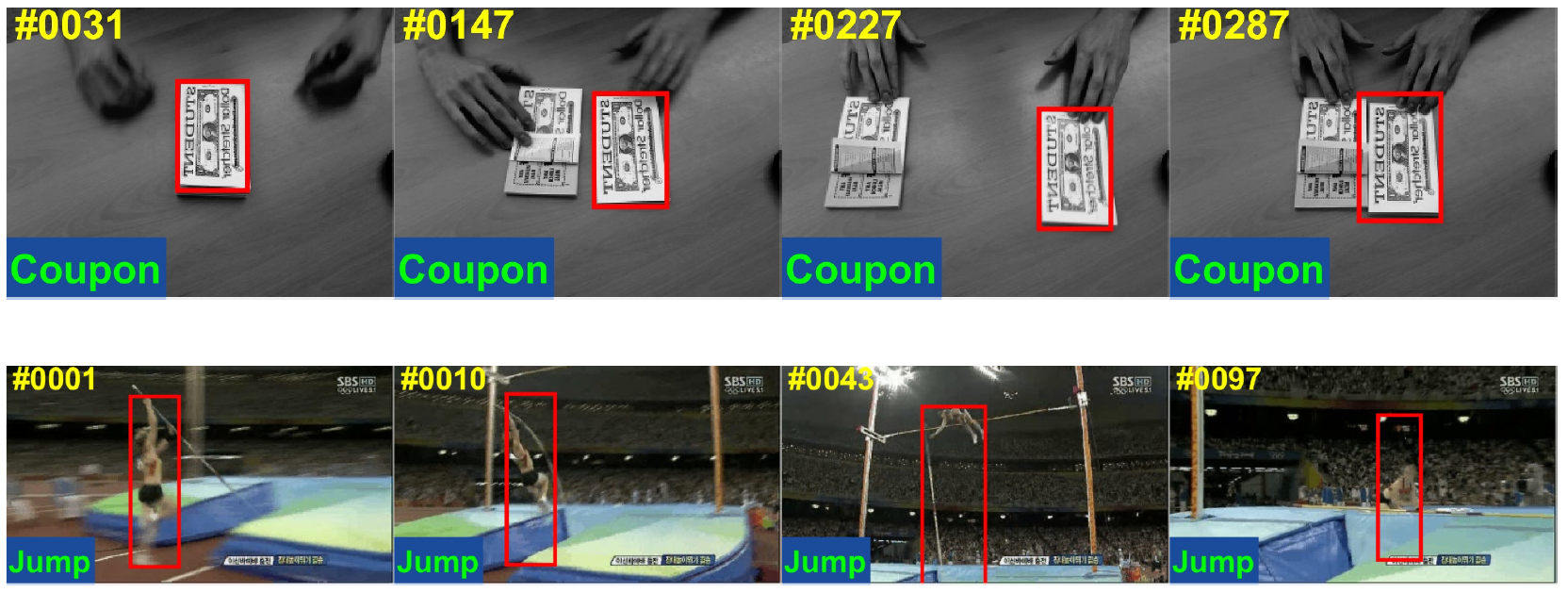}
  \caption{Failure cases on the \textit{Coupon}, and \textit{Jump} sequences. Red boxes indicate our results.}\label{fig.failure}
\end{figure}

In Figures \ref{fig.failure}, we represent two common failure cases of our algorithm, both of which result in the loss of the objects of interest. In the \textit{Coupon} sequence, our approach fails mostly because the new object that appears in the video sequence bears a striking similarity to our object of interest. In this sequence, our tracker should also tackle partial occlusion. Despite the fact that our method attempts to adjust its model based on the deformed appearance of the object of interest, it is unable to differentiate the current changed appearance from the new confounding item (distractor).
It is also worth noting that the \textit{Coupon} sequence is one of the most difficult video sequences in the OTB dataset and several state-of-the-art trackers \cite{zhu2020complementary,tian2020end,cheng2020learning} also fail to track it accurately. In addition, in the \textit{Jump} sequence, the tracker gets lost since the object movement contain a wide range of severe challenges. The movements include severe deformation, scale variation, fast motion, in-plane-rotation, out-of-plane-rotation, and occlusion challenges. A combination of the mentioned severe challenges make it hard for our tracker to track the object of interest accurately.

\section{Conclusions}\label{sec.6}
In this paper, we propose a domain adaptation approach to capture the context-aware information for the tracking-specific domain from an off-the-shelf deep model. Equipped with the proposed domain adaptation strategy, the inter- and intra-class discrepancies would be favorably increased for our visual tracker. As a result, our tracker would be able to effectively cope with occlusion and background clutter challenges. Besides, we also incorporate a cost-sensitive loss into online learning to strike a balance between positive/negative candidates and also between non- and semantic background candidates, thereby making the learning procedure unbiased during tracking procedure. Finally, experimental results on different datasets demonstrate that our approach performs satisfactorily
against the-state-of-the-art trackers in terms of accuracy and speed criteria.

\newpage
\bibliography{Manuscript_arxiv}

\end{document}